\documentclass[letterpaper, 10 pt, conference]{ieeeconf}

\IEEEoverridecommandlockouts                              
\usepackage[compatibility=false]{caption}
\usepackage{cite}
\usepackage{graphicx}
\usepackage{amsmath}
\usepackage{amssymb}
\usepackage{algorithm,tabularx}
\usepackage{mathtools,bbm}
\usepackage{array,tabularx}
\usepackage{multirow}
\usepackage[table,xcdraw,dvipsnames]{xcolor}
\newcolumntype{P}[1]{>{\centering\arraybackslash}p{#1}}
\newcolumntype{M}[1]{>{\centering\arraybackslash}m{#1}}
\usepackage{soul}
\usepackage[normalem]{ulem}
\usepackage{boldline}
\usepackage{algorithm,algorithmic}
\usepackage{adjustbox}
\usepackage{booktabs}
\usepackage{caption}
\usepackage{listings}
\usepackage{cleveref}
\usepackage{url}
\usepackage{bm}
\usepackage{mathrsfs}
\usepackage{tabularx}
\usepackage{textcomp}

\usepackage{xcolor}
\definecolor{rev_color}{RGB}{0,170,0}
\definecolor{rev_color}{RGB}{0,0,0}

\usepackage{ulem}

\usepackage{soul}
\newcommand{\soutn}[1]{\textcolor{Blue}{\st{#1}}}

\renewcommand{\soutn}[1]{\unskip}

\title{\LARGE  \bfseries
A Model Predictive Control Framework to Enhance Safety and Quality in Mobile Additive Manufacturing Systems}

\author{Yifei Li$^{1}$, Joshua A. Robbins$^{1}$, Guha Manogharan$^{1}$, Herschel C. Pangborn$^{1}$ and Ilya Kovalenko$^{2}$
\thanks{$^{2}$Yifei Li, Joshua A. Robbins, Guha Manogharan, and Herschel C. Pangborn are with the Department of Mechanical Engineering, The Pennsylvania State University, University Park, PA, USA, e-mail: \{ybl5717; jrobbins; gum53; hcpangborn\}@psu.edu.
}
\thanks{$^{2}$Ilya Kovalenko is with the  Department of Mechanical Engineering and the Department of Industrial and Manufacturing Engineering, The Pennsylvania State University, University Park, PA, USA, e-mail: iqk5135@psu.edu.}%
}

\begin{document}

\maketitle

\begin{abstract}
In recent years, the demand for customized, on-demand production has grown in the manufacturing sector. 
Additive Manufacturing (AM) has emerged as a promising technology to enhance customization capabilities, enabling greater flexibility, reduced lead times, and more efficient material usage.
However, traditional AM systems remain constrained by static setups and human worker dependencies, resulting in long lead times and limited scalability.
Mobile robots can improve the flexibility of production systems by transporting products to designated locations in a dynamic environment. 
By integrating AM systems with mobile robots, manufacturers can optimize travel time for preparatory tasks and distributed printing operations.
Mobile AM robots have been deployed for on-site production of large-scale structures, but often neglect critical print quality metrics like surface roughness.
Additionally, these systems do not have the precision necessary for producing small, intricate components.
We propose a model predictive control framework for a mobile AM platform that ensures safe navigation on the plant floor while maintaining high print quality in a dynamic environment.
Three case studies are used to test the feasibility and reliability of the proposed systems.


\end{abstract}

\section{Introduction}
\label{sec:intro}
Additive Manufacturing (AM) and Autonomous Mobile Robots (AMRs) represent two transformative technologies in smart manufacturing systems that have the potential to enable highly customized and efficient production processes. In modern manufacturing, layer-by-layer AM fabrication allows for intricate designs in customization and rapid production~\cite{das2020current}, while AMRs introduce mobility and flexibility to enable a decentralized and adaptive manufacturing workflow~\cite{fragapane2022increasing}. A trend to combine AM and AMRs is emerging as a pivotal strategy in smart manufacturing, addressing the need for resilient, scalable systems that ensure real-time responsiveness to dynamic operational demands~\cite{kovalenko2024harnessing}.

Mobile AM Robots (MAMbots) present an opportunity to improve flexibility and throughput in production environments, particularly for urgent or custom orders. Currently, MAMbots are primarily used for the production of large-scale parts~\cite{dorfler2022additive}. Compared to fixed printers, robot printing of large-scale structures can increase printing efficiency. The performance of the MAMbot platform was evaluated under varying speeds, including pre-programmed speed control, to assess its impact on AM quality and operational productivity~\cite{zhang2018large}. 
By employing multiple cooperating robots to print simultaneously, the printing time can be significantly reduced~\cite{alhijaily2023teams}. However, MAMbots are more focused on large-scale, non-delicate structures~\cite{tiryaki2019printing}, and are not used for smaller parts with stringent quality requirements~\cite{zhao2023direct}.

While MAMbots have significant potential to enhance manufacturing systems, maintaining print quality while the AM system is in motion presents significant challenges in a shop floor context.
The physical and safety-related requirements of the manufacturing environment result in MAMbot motion constraints. These include avoiding obstacles, human-occupied areas, and equipment buffer regions while adhering to speed limits within speed-restricted zones~\cite{barbieri2021uwb}.
Vibrations, uneven terrain, and dynamic environmental changes impact the precision of printed parts, often rendering them non-compliant with industrial standards. Printing tasks such as constructing support structures and infill typically demand low precision, lower material usage, and are more tolerant to movement-induced disturbances than other tasks~\cite{li2021deep}. In contrast, high-precision tasks like printing outer contours or functional components can be highly sensitive to motion variations and require greater stability~\cite{lao2021variable}. These challenges motivate the need for robot systems that dynamically align manufacturing parameters with environmental constraints to enhance productivity.

Several studies  use advanced sensing technologies to increase the reliability and accuracy of printing~\cite{liu2019image, guidetti2024force}. However, methods designed primarily for stationary setups lack the adaptability needed for mobile AM. A control strategy that dynamically adjusts the robot’s motion based on the specific demands of the printing tasks and the surrounding terrain is needed.
In~\cite{li2024mobile}, a framework combining AM and AMRs is demonstrated to enhanced productivity through simultaneous manufacturing and transportation. This control framework included a cloud-based central control layer, a system-level planning control layer, and a manufacturing unit layer. Although control layers exchange information within a modular architecture, their specialized tasks in manufacturing necessitate real-time control synchronization. This ensures interoperability across the control framework.


Model Predictive Control (MPC) is a closed-loop control strategy that uses dynamic models to predict the future behavior of a system and optimize its performance over a finite horizon~\cite{borrelli2017predictive}. This optimization is repeated with a receding horizon as time advances. Many MPC frameworks for mobile robots focus on trajectory tracking, path following, obstacle avoidance, or energy efficiency~\cite{zuo2020mpc, robbins2024energy}. MPC has also been applied in additive manufacturing for thermal management, material deposition modeling, and real-time monitoring of iterations~\cite{schmidtke2023model, inyang2020layer}.

In this work, we extend~\cite{li2024mobile} by developing an MPC framework to synchronize AM process parameters and mobile platform dynamics for MAMbots.
Specifically, the major contributions of this work are as follows:
(1) The development of predictive speed-quality models to identify potential defects in mobile additive manufacturing systems. (2) The design of a scalable MPC architecture to ensure robustness of the MAMbot platform.
(3) The integration of temporal constraints and spatial constraints for the MPC framework to increase the adaptability of MAMbots.
Three case studies show how MAMbots can ensure high print quality while moving safely despite external disturbances, including obstacles and uneven terrain.

The remainder of this manuscript is organized as follows: ~\Cref{sec:model} presents the modeling framework for MAMbot systems, consisting of a quality-speed profile and double integrator vehicle model. ~\Cref{sec:mpc} presents the MPC formulation. ~\Cref{sec:case} presents three case studies to demonstrate the proposed method. ~\Cref{sec:conclusion} provides concluding remarks.







\section{Modeling}
\label{sec:model}

\begin{figure}
\centering
\smallskip
\smallskip
\captionsetup{belowskip=-1pt}
\includegraphics[width=0.9\columnwidth]{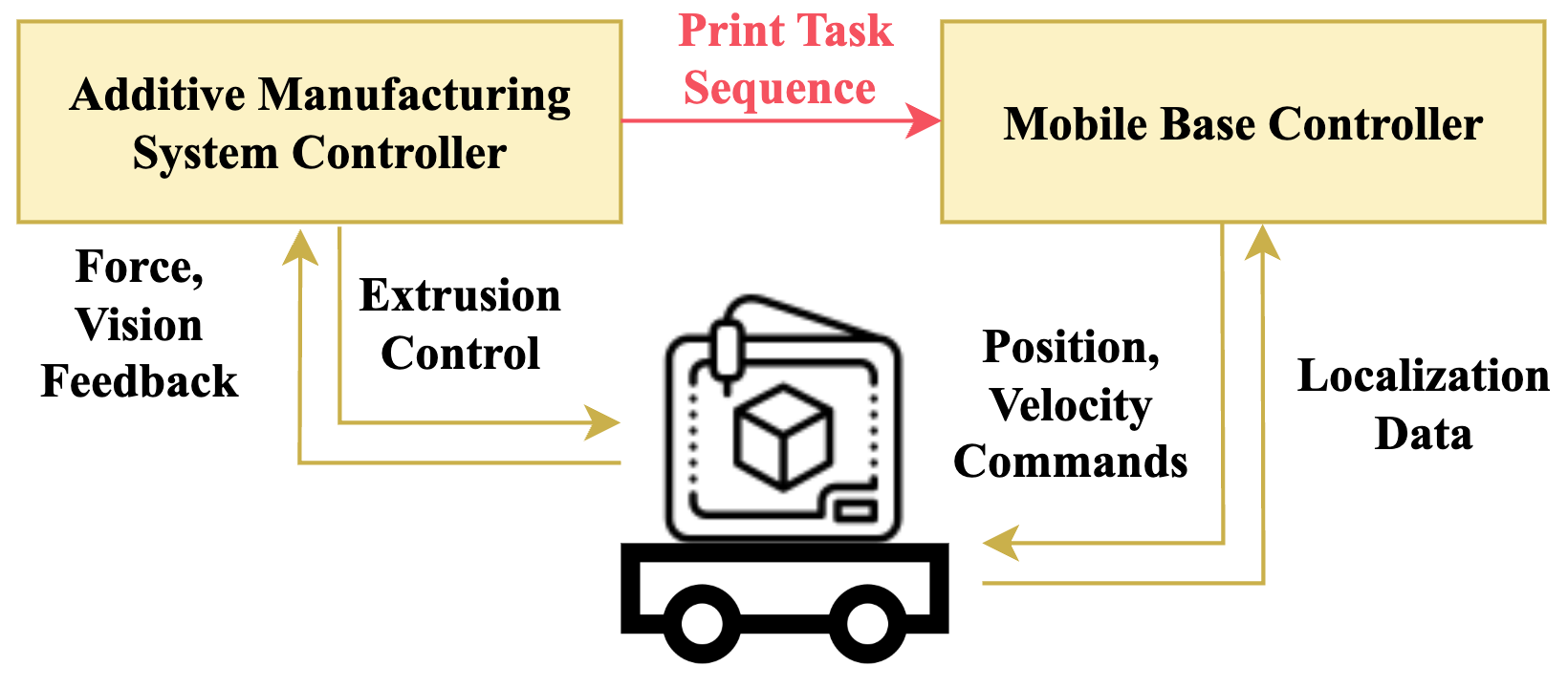}
\caption{Proposed coupled control of the AM unit and mobile unit. The AM unit shares information with the mobile base unit, e.g., print task sequence.}
\label{fig:system}
\end{figure}

This work builds on the MAMBot control architecture introduced in~\cite{li2024mobile}, with novel improvements, shown in ~\Cref{fig:system}.
The AM system controller manages the AM process by setting parameters such as extrusion speed and nozzle movement to maintain high print quality, while the mobile base controller steers the base. 
Effective coupling of the two systems is critical, as the systems necessitate adaptive adjustments to their operational coordination. 
For instance, during critical tasks such as contour printing, the platform may need to halt or decelerate to ensure dimensional accuracy and surface finish integrity.
In this work, we propose a framework for coupling the AM unit with a mobile base unit by developing constraints according to a quantitative quality-speed relationship and using an MPC framework for motion planning.


\subsection{Quality-Speed Profile} 

Traditional stationary approaches often treat quality and speed as isolated variables, relying on static parameters that do not adapt to real-time environmental or task-specific demands. 
We propose the Quality-Speed Profile (QSP) as a novel method to quantify the trade-off between manufacturing precision and robot speed in MAMbots. 

To define the QSP, we model the relationship between the mobile base speed $v$ and the predicted print quality metric $q_i(v)$, such as dimensional error. In this paper, the QSP is task-specific and constructed from experimental data using a linear fit, shown in~\Cref{fig:quality-speed}. The linear model is given by 

\begin{equation}
q_i(v) = a_i v + b_i
\label{eq:QSP}
\end{equation}
where $i$ indexes the task (e.g., shell, infill, support), and $a_i$ and $b_i$ are task-specific coefficients derived from experimental calibration. 

By defining the relationship between motion speed and achievable print quality, 
the QSP can provide recommended maximum speed references for real-time algorithms to balance production throughput with dimensional accuracy, surface finish, and structural integrity.

To create the QSP, we slice CAD models into G-Code (via slicing software) to separate tasks into contour, infill, and support tasks. The MAMbot’s system-level planner analyzes the G-Code to classify tasks by quality requirements, as shown in ~\Cref{fig:task}. For example, low-precision tasks (such as printing infills or supports) are less sensitive to minor disturbances caused by movement, allowing the robot to move at a higher speed. In contrast, high-precision tasks (such as printing outer contours or functional components) require minimal disturbances. Therefore, the robot must reduce speed to mitigate the effects of vibration and ensure stable layer deposition. The QSP establishes a predictive quality-speed relationship, enabling the controller to optimize adjustments during operation. 

\subsection{Double Integrator Vehicle Model}


We use a discrete-time Double Integrator Vehicle Model (DIVM) for MPC-based motion control of the MAMbot. This is given by
    $x_{k+1} =A x_k + B u_k \;$,
where

\begin{equation}
A =
\begin{bmatrix}
1 & 0 & \Delta t & 0 \\
0 & 1 & 0 & \Delta t \\
0 & 0 & 1 & 0 \\
0 & 0 & 0 & 1
\end{bmatrix}, \quad
B =
\begin{bmatrix}
0 & 0 \\
0 & 0 \\
\Delta t & 0 \\
0 & \Delta t
\end{bmatrix}\;,
\label{eq:AB}
\end{equation}
and $\Delta t$ is the time step interval. The DIVM is a widely-used reduced-order model in robot motion planning \cite{matni2024quantitative}. It can accurately model both systems capable of omnidirectional motion (e.g., robots with mecanum wheels) and some differentially flat non-holonomic systems such as the unicycle model~\cite{sira2018differentially}.
The DIVM represents the robot's system in terms of point mass linear motion with the state given by $\bm{x} =[x, y, \dot{x}, \dot{y}]^T$ and the control inputs given by $\bm{u} =[\ddot{x}, \ddot{y}]^T$, where $x$ and $y$ are position coordinates.

\section{MPC for MAMbot}
\label{sec:mpc}
\subsection{Speed Constraints for Print Quality}

\begin{figure}
\centering
\smallskip
\smallskip
\captionsetup{belowskip=-1pt}
\includegraphics[width=0.8\columnwidth]{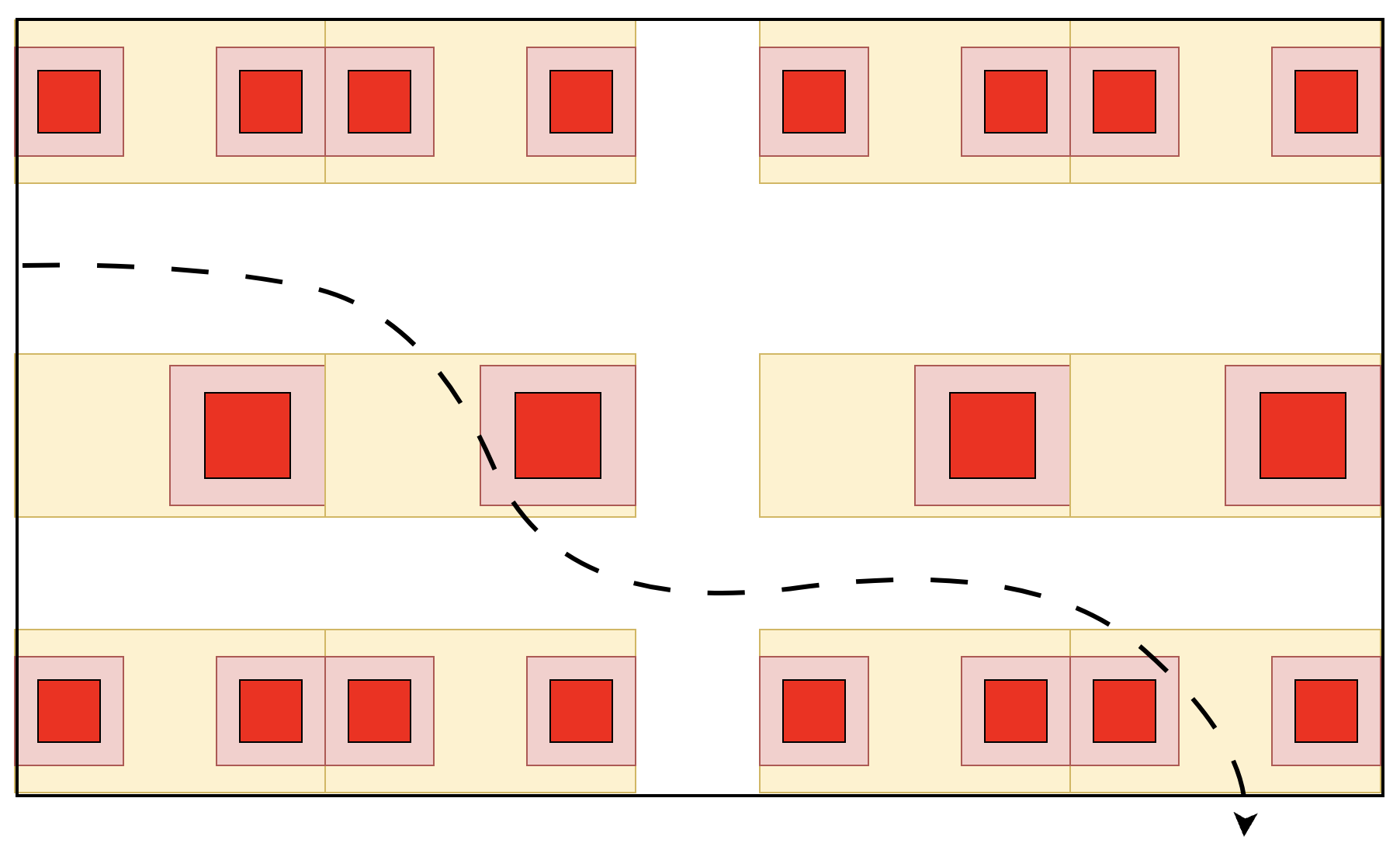}
\caption{Example overhead map of a plant floor, where red areas are obstacles, pink areas require the minimum movement speed, yellow areas require medium movement speed, and white areas have no speed limit.}
\label{fig:Terrain}
\end{figure}

To achieve reliable mobile AM on the shop floor, the QSP governs the permissible motion profiles during extrusion to preserve part quality. Additionally, shop floor constraints reflect the physical and safety limitations imposed by the dynamic manufacturing environment.
Constraints on MAMbots would include their inherent hardware limitations, as well as physical constraints, obstacle avoidance, and speed limits within designated areas.
An example shop floor that captures these constraints is shown in~\Cref{fig:Terrain}.
In this example, the shop floor incorporates predefined areas with speed limits tailored to operational priorities and environmental constraints. These areas are divided into the following categories: (1) Areas occupied by machines and obstacles (red), (2) areas with high probability of worker activity or bumps that require minimum movement speed (pink), (3) areas with medium probability of worker activity or bumps that require medium movement speed (yellow), and (4) areas with low probability of worker activity with no speed limit (white). 


\begin{figure}
\centering
\smallskip
\smallskip
\captionsetup{belowskip=-1pt}
\includegraphics[width=0.9\columnwidth]{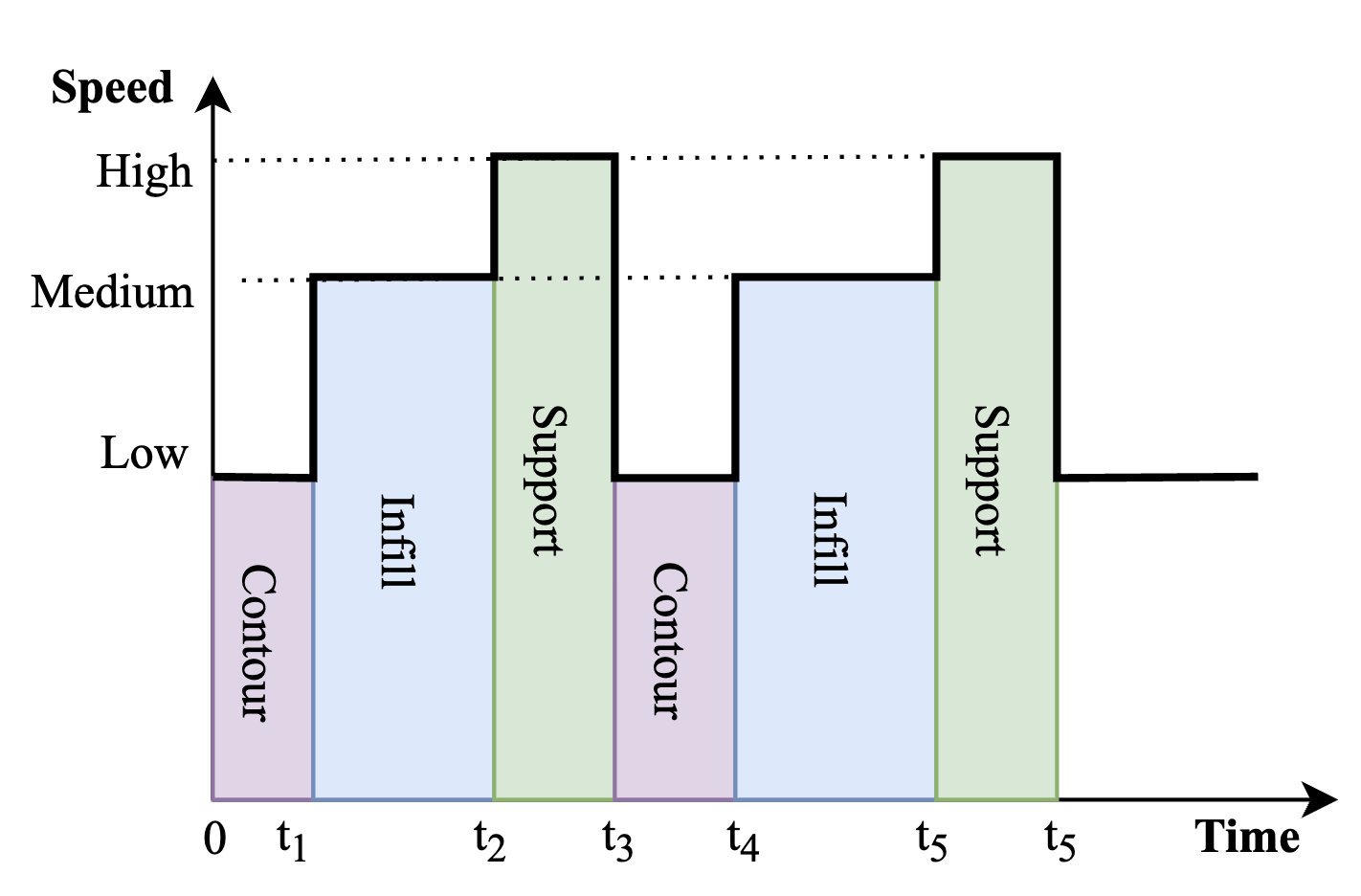}
\caption{Notional print task time sequence, which can be directly analyzed from G-code. Contours require the slowest movement speed because of their higher quality requirement. Infills and supports do not need to be of high quality, therefore the movement speed can be increased.}
\label{fig:task}
\end{figure}

The obstacle-free space is partitioned into $n_F$ convex polytopic regions $\mathcal{P}_j$, $j\in\{1,..,n_F\}$, each of which has an associated speed limit. 
In this paper, partitions are constructed via set difference. For instance, a partition $\mathcal{P}$ of the regions where minimum movement speed is required is computed as $\mathcal{P}=\mathcal{P}_{\text{min}} \setminus \mathcal{P}_{\text{obs}}$ where $\mathcal{P}_{\text{min}}$ corresponds to the pink squares in Fig.~\ref{fig:Terrain} and $\mathcal{P}_{\text{obs}}$ corresponds to the red squares. Each $\mathcal{P}_j \in \mathcal{P}$ is given in halfspace representation (H-rep) as $\mathcal{P}_j = \left\{ \begin{bmatrix} \mathbf{x} & \mathbf{y} \end{bmatrix}^T \middle| A^j_P \begin{bmatrix} \mathbf{x} & \mathbf{y} \end{bmatrix}^T \leq \mathbf{b}^j_p \right\}$.

Region-dependent velocity constraints are modeled by lifting the position constraint sets $\mathcal{P}_j$ into a higher dimensional space that includes velocity constraints. In H-rep, the coupled position and velocity constraints for a region $j$ are given as
\begin{subequations}
\begin{align}
    \mathcal{F}_j &= \left\{ \bm{x} \middle| \begin{bmatrix}
        A_p^j & 0 & 0 \\
        0 & 1 & 0 \\
        0 & -1 & 0 \\
        0 & 0 & 1 \\
        0 & 0 & -1
    \end{bmatrix} \bm{x} \leq \begin{bmatrix}
        b_p^j \\ 
        v^j_{\text{max}} \\
        v^j_{\text{max}} \\
        v^j_{\text{max}} \\
        v^j_{\text{max}}
    \end{bmatrix} \right\} \;, \\
    &= \left\{\bm{x} \middle|A^j \bm{x} \leq b^j \right\} \;,
\end{align}
\end{subequations}
where $v^j_{\text{max}}$ is the maximum allowed speed in each axis direction in region $j$.

The zone-dependent speed limit constraint can then be expressed as 
\begin{equation} \label{eq:mpc_x_in_Fk}
    \bm{x} \in \bigcup_{j=\{1,...,n_f\}} \mathcal{F}_j \;.
\end{equation}
This constraint requires that the state satisfies $\bm{x} \in \mathcal{F}_j$ for some region $j$. If $\begin{bmatrix} x & y \end{bmatrix}^T \in \mathcal{P}_j$, then \eqref{eq:mpc_x_in_Fk} implies $-v^j_{\text{max}} \leq \dot{x} \leq v^j_{\text{max}}$ and $-v^j_{\text{max}} \leq \dot{y} \leq v^j_{\text{max}}$.


Eq.~\eqref{eq:mpc_x_in_Fk} can be formulated as a mixed-integer linear constraint set using the so-called ``Big-M" method ~\cite{ioan2021mixed, hooker1994logic}. This method uses a sufficiently large scalar parameter $M \gg 1$ to embed the non-convex constraints as mixed-integer constraints. The resulting constraint set is given by
\begin{subequations}
\begin{align}
    &A^j \bm{x} \leq b^j + M\mathbf{1}(1-\zeta^j) \;, \\
    &\zeta^j \in \{0, 1\}\;, \\
    &\sum_{j=1}^{n_F} \zeta_j = 1 \;,
\end{align}
\end{subequations}
where $j \in \{1, ..., n_F\}$.

In addition, based on the analysis of the G-code file, the timing of the printing task can be obtained, as shown in ~\Cref{fig:task}. To ensure sufficient print quality, the maximum speed is limited according to the printing schedule. This results in the time-varying speed constraints $-v_{\text{max}}^{k'} \leq \dot{x}_{k'} \leq v_{\text{max}}^{k'}$ and $-v_{\text{max}}^{k'} \leq \dot{y}_{k'} \leq v_{\text{max}}^{k'}$, with
\begin{subequations}
\begin{align}
    v_{\text{max}}^{k'} &= \left\{\begin{array}{l}
    \begin{matrix}
v_{\text{contour}}, & \text{if } k'_0<k'<k'_1 \\
v_{\text{infill}}, & \text{if } k'_1<k'<k'_2 \\
v_{\text{support}}, & \text{if } k'_2<k'<k'_3 \\
...& \\
\end{matrix}
\end{array}\right. \;
\end{align}
\end{subequations}
where $v_{\text{contour}}$, $v_{\text{infill}}$ and $v_{\text{support}}$ are the speed limits for different printing tasks, and $k'_0, k'_1, k'_2, k'_3 ...$ are the changeover time steps for different printing tasks. The $k'$ notation here denotes a global time step (i.e., time steps relative to the printing task). In the MPC formulation outlined in Section~\ref{sec:MPC_form}, the global $k'$ time indices are resolved into receding horizon time indices $k$ over the moving window $k \in \{0, ..., N\}$, where $N$ is the MPC horizon length. These time-dependent state constraints are denoted as $\bm{x}_k \in \mathcal{X}_k$ for the MPC formulation in Section~\ref{sec:MPC_form}.

\subsection{MPC Formulation}
\label{sec:MPC_form}
The MPC optimization problem is defined as


\begin{subequations}  \label{eq:MPC}
\begin{align}
    &\min_{\bm{x}_k, \bm{x}_N, \bm{u}_k} \sum_{k=0}^{N-1}[(\bm{x}_k - \bm{x}_{\text{ref},k})^T Q ( \bm{x}_k - \bm{x}_{\text{ref},k}) + \bm{u}_k^T R \bm{u}_k] \nonumber\\
    &\qquad \qquad+(\bm{x}_N-\bm{x}_{\text{ref},N})^T Q (\bm{x}_N-\bm{x}_{\text{ref},N}) \;, 
    \\
    &\text{s.t.}, \; \forall k \in \{0, ..., N-1\} \;: \nonumber\\
    &\bm{x}_{k+1} =A \bm{x}_k + B \bm{u}_k \;,\; \bm{x}_{k=0} = \bm{x}_0 \;, \\
    &\bm{x}_k, \bm{x}_N \in \mathcal{X}_k \cap \left( \bigcup_{j=\{1,...,n_f\}} \mathcal{F}_j \right) \;, \label{eq:state constraints} \\
    &\bm{u}_k \in \mathcal{U} \;.
\end{align}
\end{subequations}
Eq.~\eqref{eq:MPC} is solved over a receding horizon, where the MPC discrete time horizon is $N$.
The initial condition $\bm{x}_0$ is updated at every time step using measurements from the mobile system or other sensors, and the optimal control input at time step $k=0$ ($\bm{u}_0$) is applied to the system. The input constraint set is given by $\mathcal{U}$. 
The state reference is given by $\bm{x}_{\text{ref},k}$ and assumed to be provided by a high-level path planner.
The state error term $(\bm{x}_k - \bm{x}_{\text{ref},k})^T Q ( \bm{x}_k - \bm{x}_{\text{ref},k})$ penalizes deviations in position from the state reference. The control effort term $\bm{u}_k^T R \bm{u}_k$ penalizes acceleration to promote smooth movement and stable printing. 
Equation~\eqref{eq:state constraints} enforces both the time-dependent velocity constraints and the region-dependent velocity constraints.

\section{Case Studies}
\label{sec:case}
Three case studies demonstrate the necessity and benefits of the proposed  MAMbot control framework. 
The first two of these are performed using a hardware implementation. The first case study determines how variations in movement speed influence the dimensional accuracy and surface quality of printed parts. The second case study utilizes hardware experiments to verify that varying speed control ensures high quality in additive manufacturing.
The third case study uses a simulation of a  manufacturing system to demonstrate how speed control can be applied to different areas of a shop floor.

\begin{figure}
\centering
\smallskip
\smallskip
\captionsetup{belowskip=-1pt}
\includegraphics[width=0.8\columnwidth]{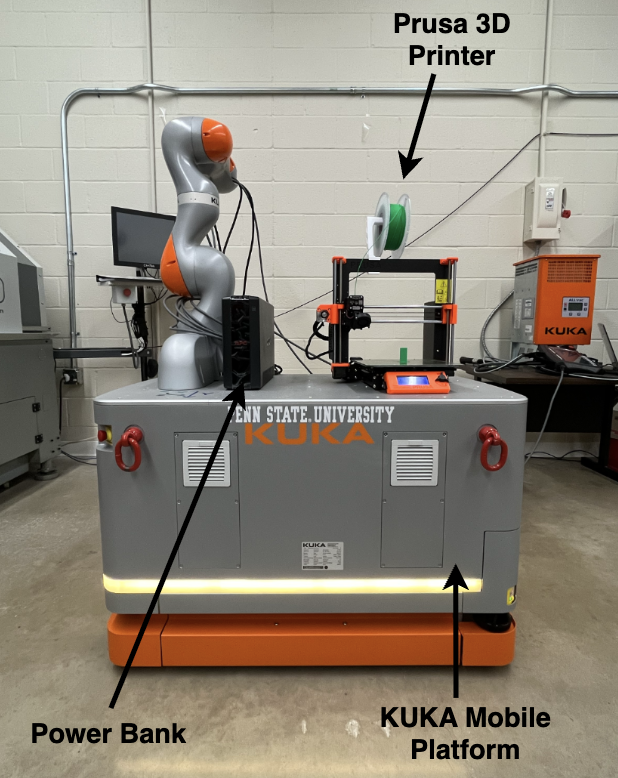}
\caption{The MAMbot platform consisting of a mobile robot, a commercial 3D printer, and a power bank.}
\label{fig:testplatform}
\end{figure}

\subsection{Hardware Implementation Case Studies} 
\label{sec:hardware}

A research platform was developed to investigate the effects of mobility on Fused Deposition Modeling (FDM) processes. The platform integrates a KUKA KMR iiwa mobile robot~\cite{KUKAWeb} controlled by the Sunrise.OS framework. The Robot Operating System (ROS)~\cite{quigley2009ros} is used for advanced motion planning and real-time communication with the mobile base~\cite{fakhruldeen2021development}. 
A commercial Prusa 3D printer is used as the AM system and attached to the KMR platform. The printer utilizes an FDM process for rapid production.
In addition, a power bank is installed to provide power to the printer and enable mobile printing capabilities, as shown in~\Cref{fig:testplatform}.

The primary objective of this MAMbot  is to fabricate a rectangular prism with dimensions of 40 mm (height) × 20 mm (width) × 2 mm (depth). 
After calibration, the printer achieved a maximum dimensional error of ±0.05 mm across all axes in static conditions, as measured by a digital caliper with 0.01 mm accuracy. For each axis, three separate measurements were taken and averaged. The reported value reflects the largest deviation from the nominal dimensions across all axes.

\subsubsection*{Case study 1} 

\begin{figure}
\centering
\smallskip
\smallskip
\captionsetup{belowskip=-1pt}
\includegraphics[width=1\columnwidth]{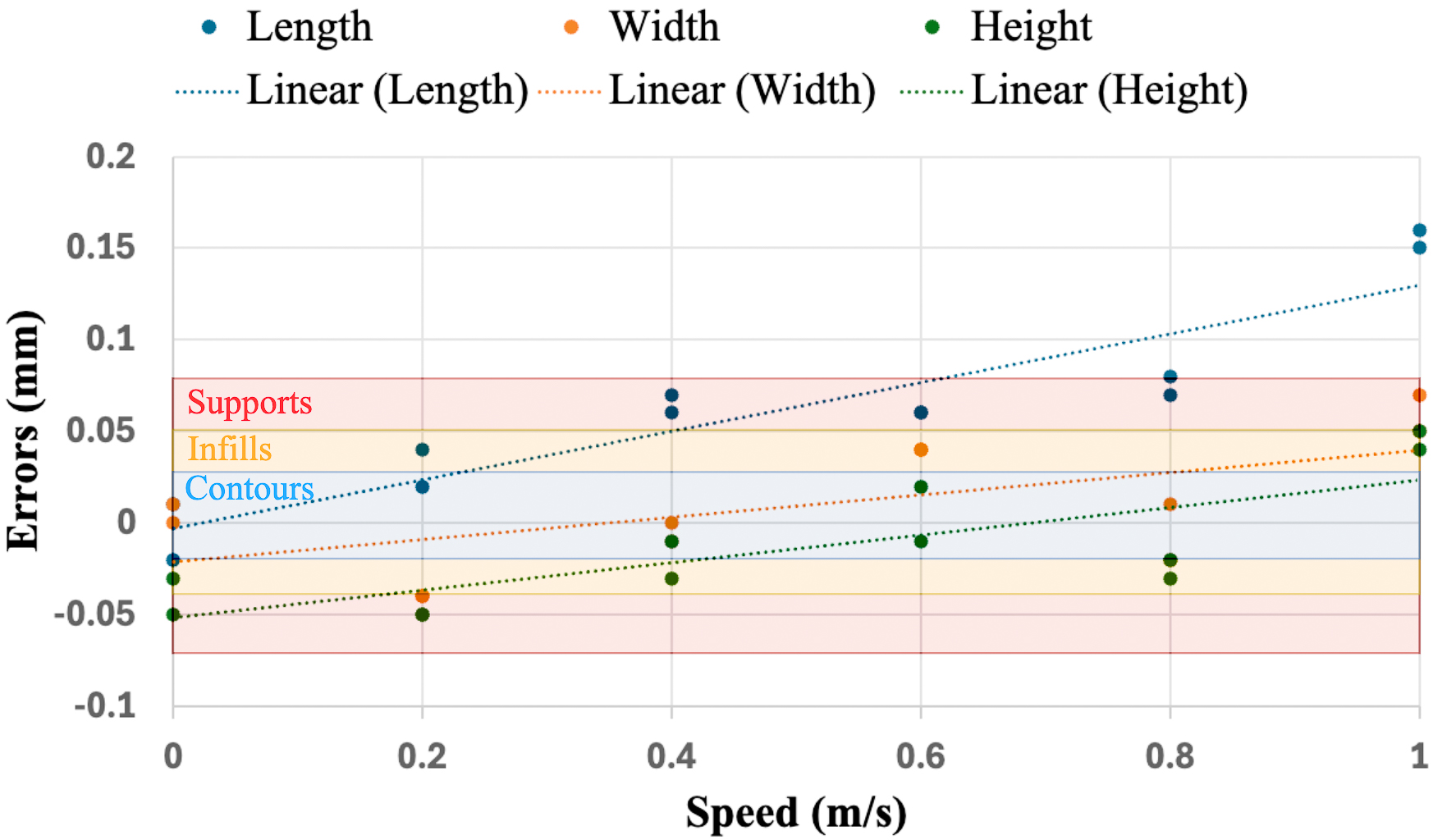}
\caption{Case study 1: Experimental data (points) and QSP (dotted lines). Blue, yellow, and red areas demarcate the acceptable error and speed tolerance ranges for contour, infill, and support structures, respectively.}
\label{fig:quality-speed}
\vspace{-0.2in}
\end{figure}

To quantify the effect of MAMbot motion on print quality, the MAMbot executed fixed-speed, straight line trajectories while printing a part. The following speeds were used: 0, 0.2, 0.4, 0.6, 0.8, and 1 m/s.
The resulting dimensional errors are shown in~\Cref{table:experiments_data}. The print quality achieved up to a speed of 0.2 m/s is comparable to stationary printing and meets the required precision tolerance range for the intended industrial applications in this paper. 
The error increases progressively with increasing speed, validating the hypothesis that vibrations from rapid movement affect additive manufacturing quality. 
Using this information, we derived an approximate QSP as shown in~\Cref{fig:quality-speed}.

\begin{table}[t]
\centering
\caption{Case study 1: Dimensional errors of printed parts at varying mobile unit speeds.}
\label{table:experiments_data}
\scriptsize
\renewcommand{\arraystretch}{1.1}
\setlength{\tabcolsep}{3pt}
\begin{tabularx}{\columnwidth}{ 
  | >{\centering\arraybackslash}X 
  | >{\centering\arraybackslash}X 
  | >{\centering\arraybackslash}X 
  | >{\centering\arraybackslash}X |}
\hline
\textbf{Speed (m/s)} & \textbf{Height (mm)} & \textbf{Width (mm)} & \textbf{Depth (mm)}  \\
\hline
\multirow{2}{*}{Stationary} & + 0.01 & 0.00 & - 0.05 \\
\cline{2-4}
& - 0.02 & + 0.01 & - 0.03 \\
\hline
\multirow{2}{*}{0.2} & + 0.04 & - 0.04 & -0.05 \\
\cline{2-4}
& + 0.02 & -0.04 & -0.05 \\
\hline
\multirow{2}{*}{0.4} & + 0.06 & - 0.01 & - 0.03 \\
\cline{2-4}
& + 0.07 & 0.00 & - 0.01 \\
\hline
\multirow{2}{*}{0.6} & + 0.06 & + 0.04 & + 0.02 \\
\cline{2-4}
& + 0.06 & + 0.04 & - 0.01 \\
\hline
\multirow{2}{*}{0.8} & + 0.08 & - 0.02 & - 0.03 \\
\cline{2-4}
& + 0.07 & + 0.01 & - 0.02 \\
\hline
\multirow{2}{*}{1.0} & + 0.16 & + 0.07 & + 0.04 \\
\cline{2-4}
& + 0.15 & + 0.05 & + 0.05 \\
\hline
\end{tabularx}
\vspace{-0.5\baselineskip}
\end{table}

\subsubsection*{Case study 2}

\begin{figure}
\centering
\smallskip
\smallskip
\captionsetup{belowskip=-1pt}
\includegraphics[width=0.8\columnwidth]{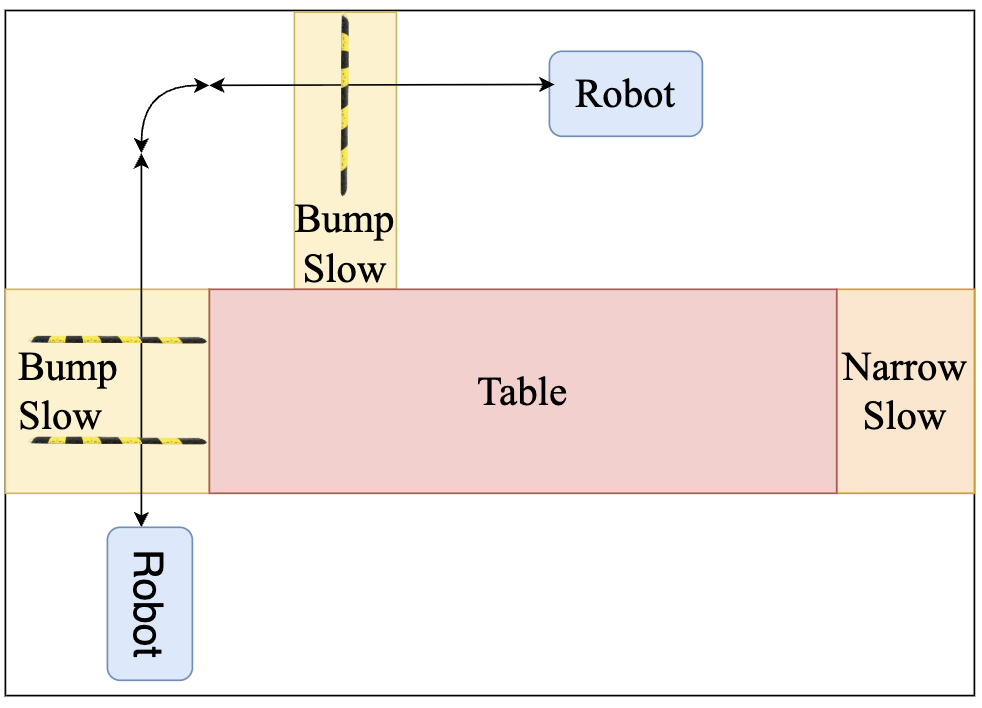}
\caption{Case study 2: Overhead view of the experimental environment, with three bumps and two slow areas.}
\label{fig:floorplan}
\end{figure}

The performance of the MAMbot platform was evaluated under varying speeds to assess its impact on AM quality and operational productivity. 
The robot was programmed to follow a predefined trajectory, starting with 3 m linear motion along the x-axis. It then executed a 90° rotation about the z-axis. Following the rotation, the robot performed an additional 3 m linear motion in the newly aligned x-direction. Finally, the platform retraced its path to complete the cycle.
The path also included three placed bumps to simulate real-world operational challenges, as shown in~\Cref{fig:floorplan}. 
The platform executed this scenario two times at fixed speeds of 0.2, 0.4, 0.6, 0.8, and 1.0 m/s, as well as using varying speed control, and the results were averaged across trials.
Under varying speed control, the speed is 0.8 m/s at maximum but reduced to 0.2 m/s when passing over bumps to maintain acceptable print quality and minimize vibration-related errors.
Each trial fabricated the same part as the first case study, with a consistent 20-minute printing time.
For this case study, the requirement for a successful print is defined as having a dimensional error within ±0.05 mm in all three directions.
The productivity (total distance traveled and cumulative turning angle) and print quality (dimensional accuracy and visual defects) of the system were measured.

\begin{figure}
\centering
\smallskip
\smallskip
\captionsetup{belowskip=-1pt}
\includegraphics[width=0.99\columnwidth]{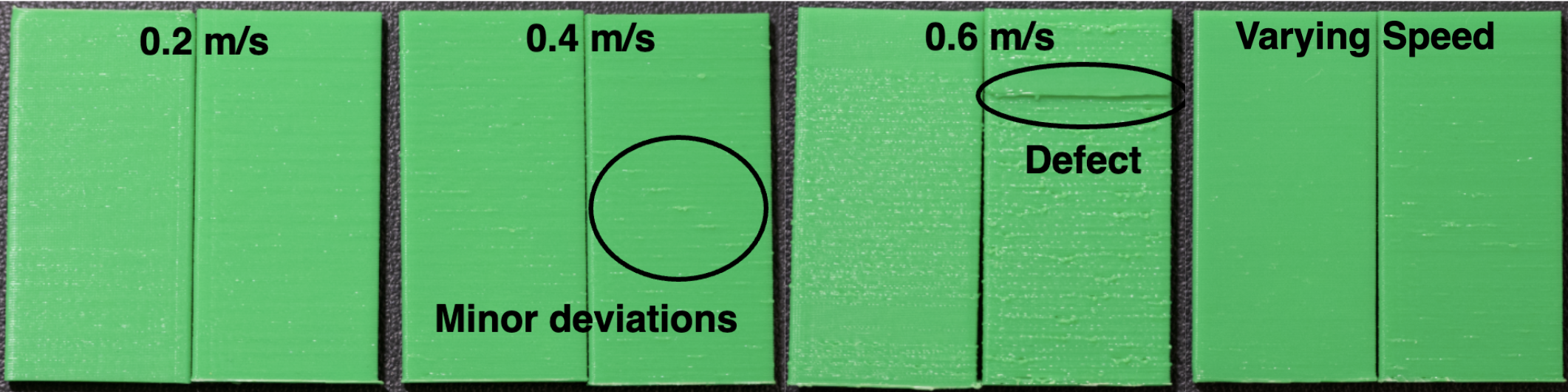}
\caption{Case study 2: Eight printed parts from experiments at 0.2 m/s, 0.4 m/s, 0.6 m/s, and varying speed control moving forward and backward during a print several times, including over bumps. The first layer is at the bottom of these images and the print direction is upward.}
\label{fig:bump}
\vspace{-0.2in}
\end{figure}

The results of these experiments are shown in~\Cref{table:experiments_data_bump} and~\Cref{fig:bump}. At the lowest speed, 0.2 m/s, the system demonstrated stable operation.
The parts met all their dimensional requirements with no observable defects. However, the conservative speed resulted in limited productivity, with a total traveled distance of 135 m and a cumulative turning angle of 69.12 rad. Increasing the speed to 0.4 m/s improved productivity (252 m traveled and turn of 131.95 rad), but introduced minor defects in print quality. No overt visual defects were detected. 
When the speed was increased to 0.6 m/s, significant quality degradation occurred, manifesting as visible layer and surface defects.
Despite these defects, productivity improvements were achieved, with the system traveling 345 m and executing a cumulative turning angle of 179.07 rad.
At 0.8 m/s and 1.0 m/s, the system experienced instability due to large vibrations. This resulted in the triggering of safety warnings and failure of the AM process. 

The varying speed strategy dynamically adjusted the platform’s speed based on floor conditions, maintaining 0.8 m/s when traversing the flat regions of the floor and reducing to 0.2 m/s in the slow areas depicted in Figure~\ref{fig:floorplan}. 
This approach balanced productivity and precision, achieving a total traveled distance of 318 m and a turning angle of 163.36 rad. These results significantly surpassed the 0.4 m/s baseline performance while avoiding the defects observed at fixed higher speeds. The printed parts exhibited no visible surface irregularities and met all dimensional tolerances. This demonstrates the efficacy of the proposed approach for mitigating vibration-induced errors during rapid motion,  
achieving better tradeoffs between throughout and quality as compared to limiting the mobile platform to a fixed speed.

\begin{table}[t]
\centering
\caption{Case study 2: Dimensional errors of printed parts at varying speeds.}
\label{table:experiments_data_bump}
\renewcommand{\arraystretch}{1.1}
\setlength{\tabcolsep}{2.5pt}
\scriptsize
\begin{tabularx}{\columnwidth}{ 
  | >{\centering\arraybackslash}X 
  | >{\centering\arraybackslash}X 
  | >{\centering\arraybackslash}X 
  | >{\centering\arraybackslash}X 
  | >{\centering\arraybackslash}X
  | >{\centering\arraybackslash}X|}
\hline
\textbf{Speed (m/s)} & \textbf{Height (mm)} & \textbf{Width (mm)} & \textbf{Depth (mm)} & \textbf{Traveled Distance (m)} & \textbf{Rotated Degrees (rad)}  \\
\hline
\multirow{2}{*}{0.2} & 0.00  & + 0.01  & - 0.03 & \multirow{2}{*}{135.00} & \multirow{2}{*}{69.12} \\
\cline{2-4}
& + 0.02  & + 0.02  & - 0.04 & & \\
\hline
\multirow{2}{*}{0.4} & - 0.05  & + 0.08  & + 0.11 & \multirow{2}{*}{252.00} & \multirow{2}{*}{131.95} \\
\cline{2-4}
& - 0.02  & + 0.09  & + 0.07 & & \\
\hline
\multirow{2}{*}{0.6} & + 0.06  & + 0.71  & + 0.27 & \multirow{2}{*}{345.00} & \multirow{2}{*}{179.07} \\
\cline{2-4}
& + 0.07  & + 0.58  & + 0.23 & & \\
\hline
0.8  & Fail  & Fail  & Fail & N/A & N/A\\
\hline
1.0  & Fail  & Fail  & Fail & N/A & N/A\\
\hline
\multirow{3}{*}{Varying} & + 0.03  & - 0.03  & - 0.05 & \multirow{3}{*}{318.00} & \multirow{3}{*}{163.36} \\
\cline{2-4}
& - 0.05  & + 0.05  & + 0.03 & & \\
\cline{2-4}
& - 0.03  & + 0.05  & 0.00 & & \\
\hline
\end{tabularx}
\vspace{-0.5\baselineskip}
\end{table}

\subsection{Factory Simulation (Case Study 3)}

Simulations were performed to evaluate the proposed approach in a larger-scale factory setting. The MPC formulation was implemented in MATLAB R2024a using the MPT Toolbox~\cite{MPT3} to compute set differences. YALMIP~\cite{lofberg2004yalmip} was used for the optimization problem formulation and the Gurobi Optimizer\cite{gurobi} was used to solve the constrained optimization problems. 

\begin{figure}
\centering
\smallskip
\smallskip
\captionsetup{belowskip=-1pt}
\includegraphics[width=1\columnwidth]{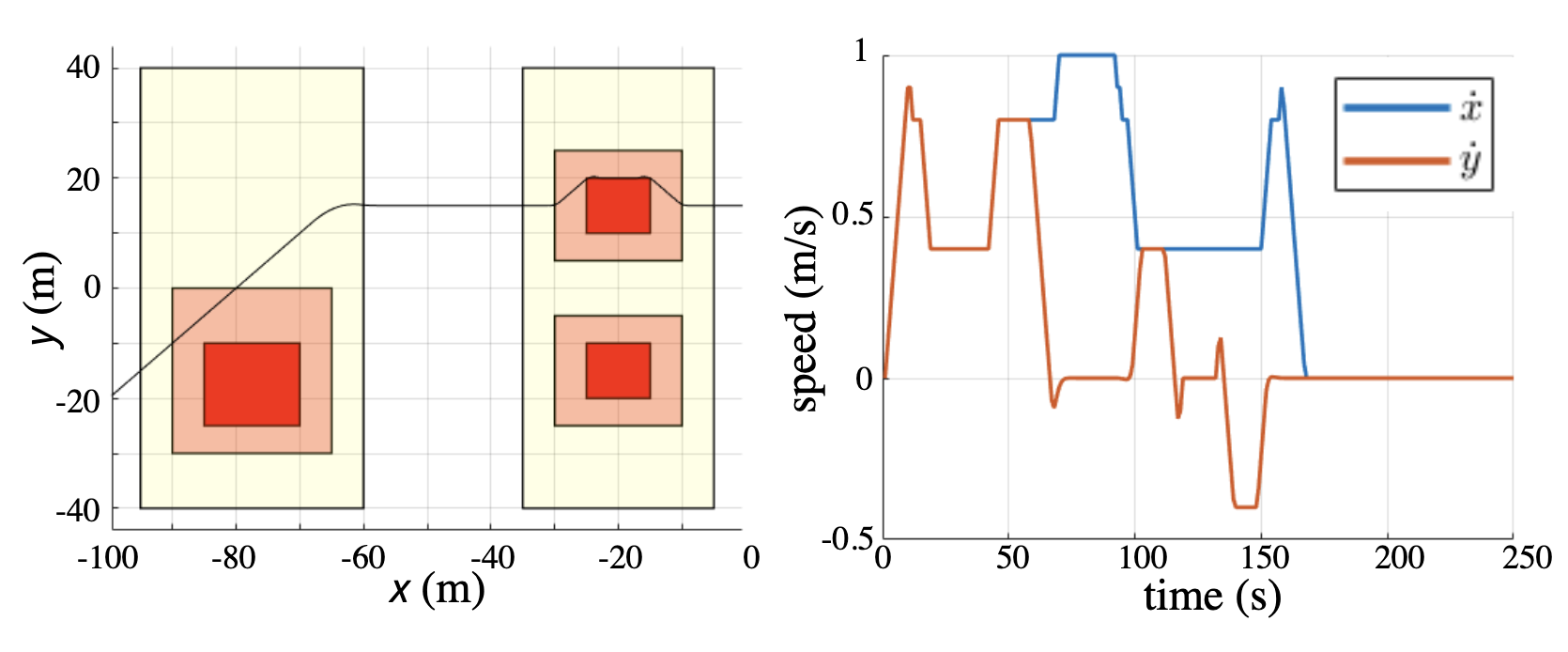}
\caption{Case study 3: Trajectory and velocity for scenario 1. The goal of the controller is to arrive at the desired destination while complying with spatially-dependent speed limits.}
\label{fig:result1}
\end{figure}

\begin{figure}
\centering
\smallskip
\smallskip
\captionsetup{belowskip=-1pt}
\includegraphics[width=1\columnwidth]{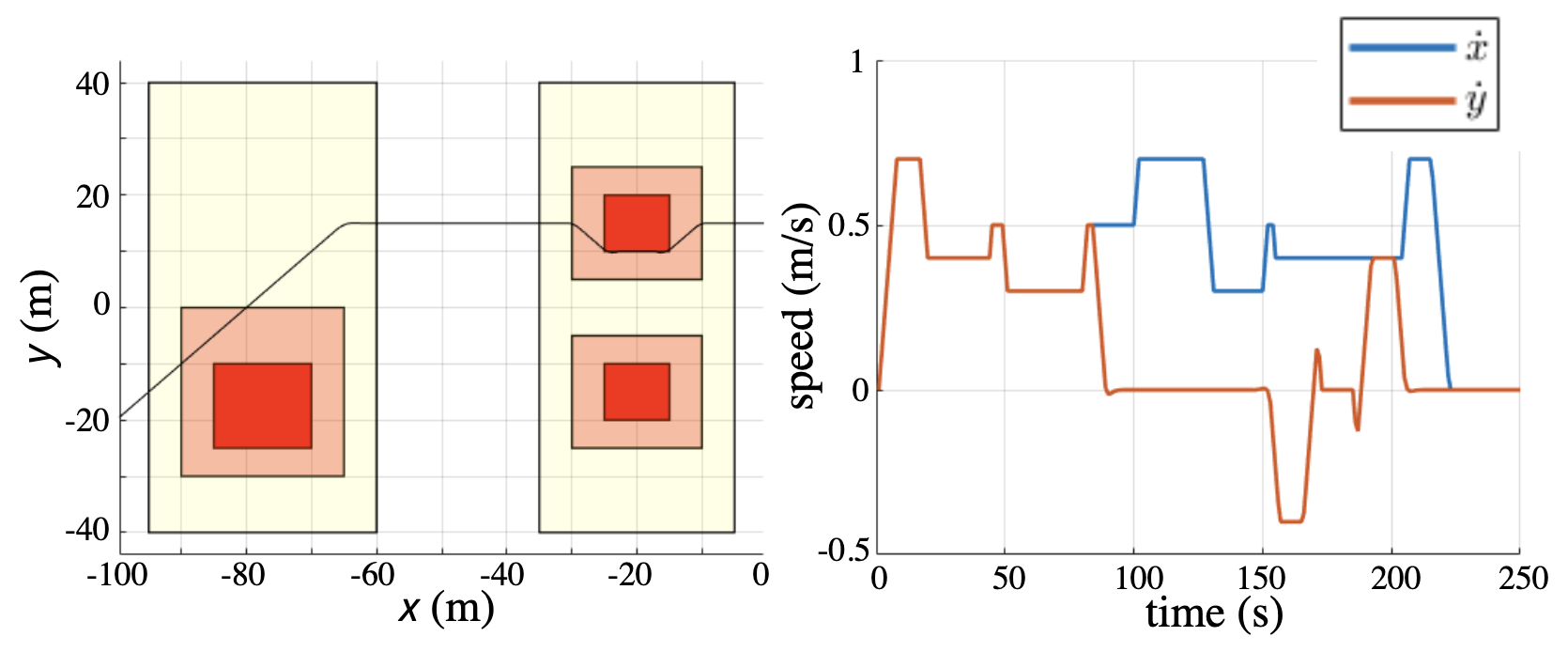}
\caption{Case study 3: Trajectory and velocity for scenario 2. The goal of the controller is to  arrive at the desired destination while complying with spatially-dependent speed limits \textit{and} constraints based on the print task time sequence.}
\label{fig:result2}
\end{figure}

Time-dependent speed constraints associated with the print time task sequence were obtained from analysis of the G-code for the rectangular prism used in the previous case studies. 
The robot moved from the initial point ($-$100, $-$20) to the end point (0, 15), as shown in the left side of~\Cref{fig:result1} and~\Cref{fig:result2}. The controller operated with a time step of 1 s and prediction horizon of 25 steps, with the simulation performed for a total of 250 s. The cost function used identity matrices for both $Q$ and $R$. The reference state was set to $\mathbf{x}_{\text{ref},k}=\begin{bmatrix} 0 & 15 & 0 & 0\end{bmatrix}^T ,\; \forall k \in \{0, ..., N \}$.
Two scenarios were used to evaluate the performance of the MPC framework, as follows:
\begin{itemize}
    \item \textbf{Scenario 1}: Spatial speed limits for different sections of the shop floor, but no constraints based on the print task time sequence.
    \item \textbf{Scenario 2}: Both spatial speed limits for different sections of the shop floor \textit{and} constraints based on the print task time sequence.
\end{itemize}

Pink/yellow regions in \Cref{fig:result1} and~\Cref{fig:result2} represent speed-restricted areas, while yellow regions represent processing machines and are not accessible to the MAMbot. While these regions remained static in this case study, future work could incorporate dynamic mapping based on worker presence probability data to ensure safe human-robot coexistence by adaptively lowering robot speed in high-activity areas.
Speed allowances for the platform scale with distance from the processing station, ensuring safe separation and operational stability. 

The following constraints were used for both scenarios:

\begin{itemize}
    \item The robot must avoid the red areas.
    \item The pink areas require that the robot's $x$ and $y$ speeds be a maximum of 0.4 m/s: $\dot{x}, \dot{y} \in [-0.4, 0.4]$.
    \item The yellow areas require that the robot's $x$ and $y$ speeds be a maximum of 0.8 m/s: $\dot{x}, \dot{y} \in [-0.8, 0.8].$
    \item The white areas allows the robot to move at full speed (1 m/s): $\dot{x}, \dot{y} \in [-1, 1]$.
\end{itemize}

In both scenarios, the MAMbot successfully navigated to its target location while satisfying constraints. 
In scenario 1, shown in~\Cref{fig:result1}, the proposed MPC framework ensured compliance with constraints by adjusting the MAMbot's speed before entering or exiting each speed-limited area. The MAMbot achieved the target position at 160 s of the simulation, demonstrating successful navigation.

The purpose of scenario 2 is to evaluate if the proposed approach can satisfy both the speed-respected area constraints and task-specific AM processing constraints. Each area has a predefined speed limit, and the robot's speed is dynamically adjusted according to the print task time sequence.
As part of the AM tasks, the contours, infills and supports are printed alternately for each layer. When the MAMbot receives a printing task, the manufacturing unit analyzes the G-code to determine print quality requirements and establish the temporal sequence for different tasks, as exemplified in~\Cref{fig:task}. 
This ensures high overall print quality and reduces movement time to increase productivity. 

The following constraints were used for scenario~2, in addition to those from scenario 1:

\begin{itemize}
    \item Contours require that the robot's $x$ and $y$ speeds be a maximum of 0.3 m/s: $\dot{x}, \dot{y} \in  [-0.3, 0.3]$ in time steps $k \in [51, 80] \cup[131,150];$
    \item Infills require that the robot's $x$ and $y$ speeds be a maximum of 0.5 m/s: $\dot{x}, \dot{y} \in  [-0.5, 0.5]$ in time steps $k \in [21,50] \cup [81,100] \cup [151,180];$
    \item Supports require that the robot's $x$ and $y$ speeds be a maximum of 0.7 m/s: $\dot{x}, \dot{y} \in  [-0.7, 0.7]$ in time steps $k \in [0,20] \cup [101,130] \cup [181,250];$
\end{itemize}

The results of scenario 2 are shown in~\Cref{fig:result2}. 
The MPC adjusted its speed to meet print quality requirements. This included reducing speeds during high-precision task to ensure optimal material placement. 
This extended the total travel time as compared to scenario 1, with the MAMbot reaching the target at 223 s. The MAMbot was sometimes moving at low speeds in areas with high speed limits because it was doing high-quality print work at the time.
For instance, at 10 s, the robot navigated an area with a 1 m/s speed limit but operated at 0.3 m/s to ensure high-quality contour deposition. This adjustment increased the travel time but ensured compliance with print quality standards.




\section{Conclusions}
\label{sec:conclusion}
Manufacturers face a critical need to customize production while meeting manufacturing deadlines.
This work advances the MAMbot framework as a solution for these dynamic industrial challenges.
An MPC framework is proposed that leverages print quality evaluation to ensure safe robot movement and high-quality printing in factory environments. 
Experimental results show that the proposed MPC framework and MAMbot system maintain the required print integrity while navigating the complex plant floor environment. Furthermore, the integration of real-time predictive speed control with AM commands demonstrates the framework's scalability for large-scale factory deployments.

Future work will integrate holistic performance metrics into the MPC framework. These metrics will include an overall print quality index and a throughput efficiency score to quantify system performance. We will also consider the computational requirements for real-time implementation of MPC.
By adding these features, we hope to improve the performance of the control architecture between robotic mobility and AM process. The framework will also be further tested in real-word manufacturing environments.

\bibliographystyle{IEEEtran}

\bibliography{yifeiBib}








\end{document}